\documentclass[letterpaper, 10 pt, conference]{ieeeconf}

\IEEEoverridecommandlockouts
\usepackage{graphicx}
\usepackage{tikz}
\usepackage{amsmath}
\usepackage{amssymb}
\usepackage{algorithm}
\usepackage{algorithmic}
\usepackage{tabularx}
\usepackage{float}
\usepackage{fvextra}
\usepackage{xcolor}
\makeatletter
\let\NAT@parse\undefined
\makeatother
\usepackage[numbers,sort&compress]{natbib}
\usepackage{listings}
\usepackage{xcolor}
\usetikzlibrary{arrows.meta,positioning}

\usepackage[hidelinks]{hyperref}

\lstdefinestyle{subgoaljson}{
  basicstyle=\ttfamily\footnotesize,
  backgroundcolor=\color[RGB]{242,242,236},
  frame=single,
  rulecolor=\color{black},
  framerule=0.4pt,
  numbers=left,
  numberstyle=\tiny\color{gray},
  numbersep=6pt,
  stepnumber=1,
  xleftmargin=1.6em,
  breaklines=true,
  breakatwhitespace=true,
  breakindent=1.5em,
  postbreak={},
  columns=fullflexible,
  keepspaces=true,
  showstringspaces=false,
  aboveskip=2pt,
  belowskip=2pt,
}

\title{\LARGE \bf
Navi-Agent: Unlocalized Monocular Navigation Agent
}

\author{
{Wenyuan Xie}$^{1*}$, {Mengyang Hong}$^{1*}$, {Yongzhong Wang}$^{2*}$, {Yanbiao Ji}$^{1}$, {Yijin Zhou}$^{1}$, {Shaokai Wu}$^{1}$, \\
{Shalayiding Sirejiding}$^{3}$, {Huayi Zhou}$^{4}$, {Yi-Chao Chen}$^{1}$, {Ma Ling}$^{1}$, {Yue Ding}$^{1,\dagger}$, {Hongtao Lu}$^{1,\dagger}$ \\
\vspace{0.15cm}
\thanks{$^{*}$Equal contribution.}
\thanks{$^{\dagger}$Corresponding authors.}
\thanks{$^{1}${Shanghai Jiao Tong University}.}%
\thanks{$^{2}${Southern University of Science and Technology}.}%
\thanks{$^{3}${Xinjiang University of Finance and Economics}.}%
\thanks{$^{4}${Shenzhen University}.}%
}

\begin{document}
\maketitle
\thispagestyle{empty}
\pagestyle{empty}


\begin{abstract}
Vision-Language Navigation in Continuous Environments (VLN-CE) requires an embodied agent to execute long-horizon instructions in unknown environments. Existing zero-shot VLN-CE systems typically maintain spatial states through geometric localization or coordinate-based representations. Recent geometry-constrained navigation removes depth and globally consistent coordinates, but maintaining persistent spatial awareness for place confirmation, progress verification, and recovery remains challenging. We present Navi-Agent, a zero-shot VLN-CE agent that constructs a coordinate-free spatial state from visual observations and executed motion histories. Navi-Agent organizes this state as a navigation topology, where nodes represent visual places and edges represent motion transitions. This representation enables observation-based approximate self-localization, task progress verification, and visual revisitation-based recovery. Navi-Agent performs closed-loop navigation by decomposing instructions into sub-goals, executing local visual navigation, and verifying visited places through the constructed spatial state. Experiments on zero-shot VLN-CE benchmark and real-world robot platforms show that Navi-Agent achieves state-of-the-art performance among geometry-constrained methods while remaining competitive with approaches relying on geometric localization.
\end{abstract}

\section{Introduction}

\begin{figure}[t]
    \centering
    \includegraphics[width=\linewidth]{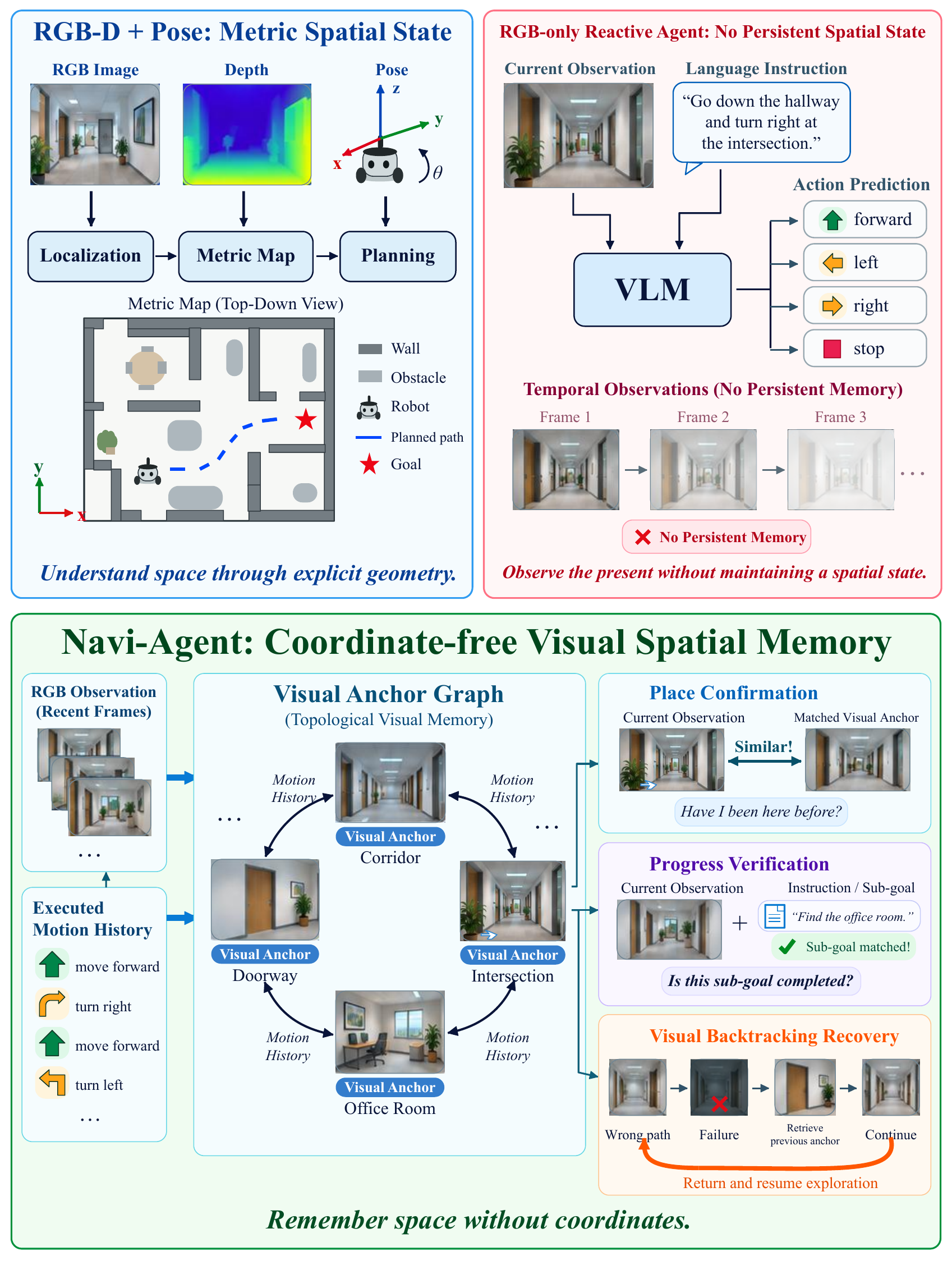}
    \vspace{-20pt}
    \caption{Navi-Agent overview. The proposed agent builds a coordinate-free Visual Anchor Graph to maintain visual spatial memory for long-horizon navigation.}
    \label{fig:teaser}
    \vspace{-20pt}
\end{figure}

With the development of large-scale vision-language foundation models~\citep{radford2021learning,achiam2023gpt}, zero-shot Vision-and-Language Navigation in Continuous Environments (VLN-CE) has become increasingly practical~\citep{krantz2020beyond,hong2022bridging}. VLN-CE requires an embodied agent to follow natural-language instructions in an unknown environment without task-specific expert trajectories. During navigation, the agent needs to understand instructions, perceive the environment, and select actions based on continuous observations. Existing zero-shot VLN-CE methods have gradually adopted a common system paradigm: decomposing long instructions into ordered sub-tasks, executing local actions for each sub-task, and using subsequent observations to estimate progress, update memory, and determine stage transitions~\citep{zhou2024navgpt,chen2024mapgpt,long2024instructnav,qiao2025open,chen2025constraint}. A key component of these methods is maintaining a spatial representation that associates observations across different time steps, including explicit BEV or voxel maps~\citep{an2022bevbert,huang2023visual,chen2025constraint},  3D scene graphs~\citep{gu2024conceptgraphs} or landmarks graphs~\citep{shah2023lm,yin2025gc}.

However, the construction of these spatial representations relies on global coordinates obtained either directly or indirectly. To remove this dependence on geometric localization, recent studies have explored geometry-constrained navigation, where depth information and globally consistent coordinates are unavailable~\citep{wang2025dreamnav,luo2026lightzeronav}. Within this restriction, existing work mainly compensates for missing global spatial relations in two ways. One line uses local relative trajectories or predicted future observations to compare an executed action with its possible visual outcome~\citep{wang2025dreamnav}. The other retains short-term observation histories or stage-level visual references, allowing a vision-language model to estimate sub-task progress through observation matching and semantic judgment~\citep{luo2026lightzeronav}. These methods show that local visual evidence can support action selection and stage progression to some extent. Specifically, the agent may have difficulty determining whether the current observation corresponds to a previously visited place, verifying whether an executed motion has completed the intended transition, and recovering to a reliable location after deviating from the instruction. This leads to the central question of this work: how can an agent maintain a persistent spatial state from visual observations without explicit geometric localization, verify its current place and task progress, and recover from navigation failures through visual revisitation?

To address this question, we present Navi-Agent, a zero-shot navigation agent for a strictly visual setting. To avoid reliance on explicit geometric localization, Navi-Agent transforms the conventional metric-localization-based navigation topology into a navigation history topology grounded in visual observations and executed motions. In this topology, nodes are represented by observed visual scenes and edges are represented by executed motion histories, enabling long-horizon navigation without explicit geometric states.
During navigation, Navi-Agent follows the standard long-horizon navigation pipeline of instruction decomposition, local target selection, and short-range execution. For each active sub-task, the agent selects a local visual target and performs navigation using a low-level controller. After reaching the local target, the agent records the visual observation and semantic context of the current scene as a topology node and stores the executed motion history from the previous node as a topology edge. The agent then determines whether the current target is completed based on the current observation, sub-instruction, and navigation history. If completed, it proceeds to the next sub-task; otherwise, it continues exploring and verifying alternative visual targets for the current sub-task.
When exploration fails or the current branch is determined to be incorrect, the agent retrieves historical visual nodes, follows the history keyframe pointtracking for backtracking, and uses visual matching to confirm revisitation before exploring alternative directions. The entire process requires no depth, absolute pose, odometry, or privileged simulator geometry, and enables an observation-driven navigation loop for spatial state maintenance, task progression, and error recovery.

Our contributions are summarized as follows:
\begin{itemize}

\item To the best of our knowledge, Navi-Agent is the first framework that explores spatial state construction for zero-shot VLN-CE under geometry-constrained settings, enabling observation-based approximate self-localization without explicit spatial coordinates.

\item We design a coordinate-free navigation topology representation as an alternative to metric-localization-based spatial topology. The proposed representation defines nodes from visual scenes and edges from executed motion histories, enabling spatial state maintenance and task progress verification without explicit geometric states.

\item We propose a visual revisitation-based closed-loop recovery mechanism that integrates historical visual state retrieval, motion-history-based backtracking, and visual matching confirmation for robust long-horizon navigation.

\end{itemize}


\begin{figure*}[!t]
    \centering
    \includegraphics[width=0.92\textwidth]{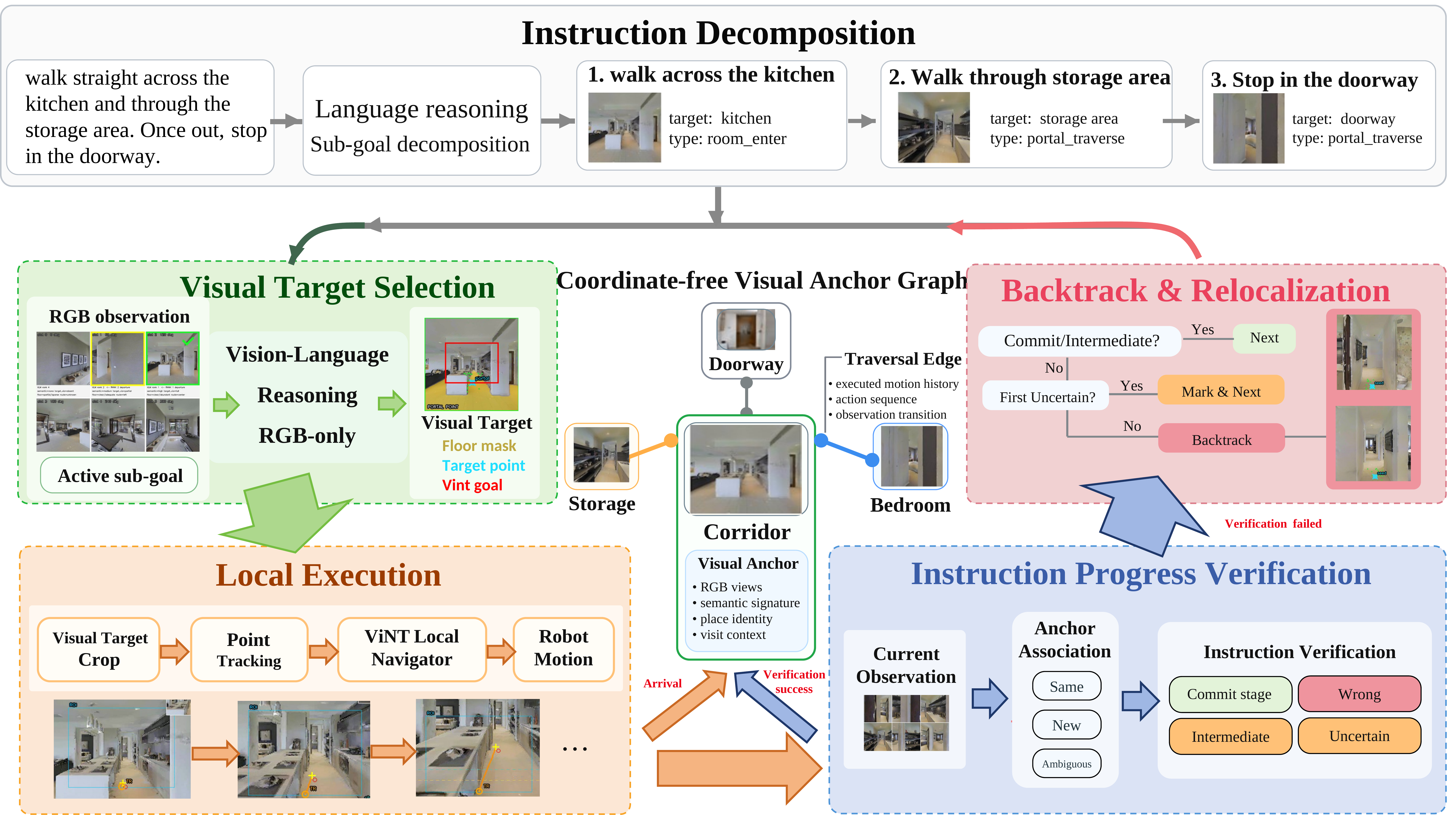}
    \caption{Navi-Agent system overview. A long-horizon instruction is first decomposed into ordered typed sub-goals (\textit{Decompose}). For each active sub-goal, \textit{SelectTarget} ranks six RGB views with a VLM and grounds the selected visual target (floor mask, trackable point, and ViNT goal crop); \textit{ExecuteLocal} then performs navigation via point tracking and ViNT. Each physical stop is registered into a coordinate-free visual anchor graph, where nodes store RGB views, semantic signatures, place identities, and visit contexts, and edges record executed motion. \textit{Associate} identifies each new visit as \texttt{Same}/\texttt{New}/\texttt{Ambiguous}, while \textit{VerifyProgress} determines stage transition as \texttt{Commit}/\texttt{Intermediate}/\texttt{Wrong}/\texttt{Uncertain}. Upon verification failure, the agent explores alternatives or backtracks through recorded edges to re-confirm the parent place. }
    \label{fig_method}
\end{figure*}

\section{Related Work}
\label{sec:related_work}

Zero-shot VLN-CE agents typically organize long instructions into ordered goals and alternate between instruction understanding, local execution, progress estimation, and action decision-making over continuous observations. NavGPT converts visual observations into language descriptions for a GPT-style model to select candidate directions; DiscussNav separates instruction analysis, scene perception, and completion estimation through multi-expert discussion; MapGPT adds online visited nodes and their topological relations to the prompt for multi-step planning and branch exploration; and Open-Nav combines spatiotemporal chain reasoning, waypoint selection, and local execution in continuous environments~\citep{zhou2024navgpt,long2024discuss,chen2024mapgpt,qiao2025open}. These works establish a common agentic paradigm of instruction analysis, stage-wise execution, and completion judgment~\citep{long2024instructnav}. Their main differences lie in how spatial evidence is provided for progress estimation and action execution.

Existing navigation systems organize spatial state through metric maps~\citep{an2022bevbert,huang2023visual}, topological graphs~\citep{an2024etpnav,chen2024mapgpt}, and semantic memories. ETPNav, waypoint-based models, and history-modeling methods align observations across time using depth, camera pose, or odometry to support obstacle modeling, distance computation, and path planning~\citep{an2024etpnav,krantz2021waypoint,chen2021history,yu2025mossvln}. Semantic maps and scene graphs in ObjectNav further store objects, landmarks, and room relations in retrievable representations for target search and relocation~\citep{yokoyama2024vlfm}. Correspondingly, task completion is judged through waypoint arrival, distance or occupancy states, sub-instruction constraints, textual history, or visual similarity. Zero-shot systems such as CA-Nav, MVP-Nav, and SmartWay ground language decisions in continuous actions through value maps, constraint states, depth-enhanced waypoints, and historical backtracking~\citep{chen2025constraint,xie2026mvp,shi2025smartway,shi2025fast,yin2025gc}. These representations provide relatively stable spatial scaffolds for long-horizon navigation, while task confirmation relies on evidence from distance, constraints, visual similarity, or historical information.

Recent studies reveal a semantic-to-geometric gap in VLM spatial reasoning: VLMs excel at qualitative semantic inference but are unreliable when geometric relations must be derived implicitly. GCA addresses this by confining the VLM to translating queries into formal, verifiable task constraints and delegating computation to geometric tools~\citep{chen2026geometrically}. This line constrains reasoning \emph{with} geometry, whereas geometry-constrained navigation, as studied here, restricts the geometric sensing available to the agent. To improve cross-platform generalization and probe the spatial reasoning capabilities of foundation models themselves, a small number of recent zero-shot methods have begun to reduce their dependence on global metric state. DreamNav uses local relative trajectories and future visual prediction to evaluate candidate paths through imagined outcomes; LightZeroNav retains short-term trajectory observations and stage-level visual baselines, and estimates progress through image-level comparison before switching sub-goals~\citep{wang2025dreamnav,luo2026lightzeronav}. These methods represent predictive and observation-comparison routes within geometry-constrained navigation, showing that local visual and relative-motion cues can support action selection and stage progression. However, without a globally consistent coordinate relation, how to organize historical observations into persistent place identities and spatial transitions for location confirmation, task progression, and error recovery remains insufficiently addressed.



	


\section{Methodology}
\label{sec:methodology}

Navi-Agent organizes long-horizon navigation as a sequence of verifiable local loops. Instead of directly predicting actions with a VLM (Vision-Language Model), the agent identifies the active sub-goal, selects an executable visual target, performs a point-tracking-guided motion segment, and verifies the visited place against the instruction. Only a verified stage advances the instruction; otherwise, the observation is retained for exploration or
backtracking.

\subsection{Problem Statement}
\label{subsec:method_problem}

We study zero-shot VLN-CE~\citep{krantz2020beyond} in a geometry-constrained indoor environment using only a single forward-facing RGB camera, where the agent starts from an episode-specific pose and navigates to the goal described by a natural-language instruction $I$ without depth or metric coordinates. At time $t$, given observation $o_t$, the policy selects $a_t=\pi(o_t,I,\mathcal{M}_t)\in{Action Set}=\{\texttt{move\_forward},\texttt{turn\_left},\texttt{turn\_right},\texttt{stop}\}$, leveraging a memory graph $\mathcal{M}_t=(\mathcal{A}_t,\mathcal{E}_t)$ of visual anchor nodes $\mathcal{A}_t$ and executed motion edges $\mathcal{E}_t$ (Sec.~\ref{subsec:method_memory}) to successfully issue \texttt{stop} within the target distance threshold.

\textbf{Implementation Details:} Our setting only receive RGB observations from the environment. Other information like Depth, absolute pose, odometry, SLAM/SfM coordinates, and global metric maps are unavailable.
System components are built upon off-the-shelf modules without task-specific fine-tuning. We use GroundingDINO~\citep{liu2024grounding} and SAM~\citep{kirillov2023segment} for RGB target localization, ViNT~\citep{shah2023vint} for local navigation, and KLT tracking~\citep{tomasi1991detection,shi1994good} for visual target following.

\subsection{Pipeline Overview}
\label{subsec:method_pipeline}

\begin{figure*}[t]
\centering
\begin{tikzpicture}[
  font=\scriptsize,
  box/.style={draw=black!70, rounded corners=2pt, fill=black!4,
              inner sep=4pt, anchor=west, align=left,
              text width=\dimexpr0.25\textwidth-0.34in\relax,
              minimum height=0.9in},
  sub/.style={box, fill=blue!4},
  arr/.style={-{Latex[length=1.8mm]}, thick, black!70}
]
\node[box] (I) {%
  \textbf{Instruction $I$}\\[1pt]
  ``Turn left and walk straight across the kitchen and through the
  storage area. Once out, turn right and stop at doorway.''};

\node[sub, right=0.42in of I] (g0) {%
  \textbf{g0}\hfill\texttt{room\_enter}\\[1pt]
  ``Turn left and walk straight across the kitchen''\\[1pt]
  \texttt{entities}: kitchen;\ \texttt{direction}: left\\
  \texttt{query}: ``kitchen entrance doorway''};

\node[sub, right=0.15in of g0] (g1) {%
  \textbf{g1}\hfill\texttt{portal\_traverse}\\[1pt]
  ``Walk through the storage area''\\[1pt]
  \texttt{entities}: storage area;\ \texttt{direction}: straight\\
  \texttt{query}: ``storage area entrance''};

\node[sub, right=0.15in of g1] (g2) {%
  \textbf{g2} (terminal)\hfill\texttt{portal\_stop}\\[1pt]
  ``Turn right and stop at doorway''\\[1pt]
  \texttt{entities}: doorway \\
  \texttt{relation}: on the right;\ \texttt{ordinal}: last\\
  \texttt{query}: ``doorway on the right''};

\draw[arr] (I) -- node[above, font=\tiny, inner sep=1pt] {\textit{Decompose}} (g0);
\draw[arr] (g0) -- (g1);
\draw[arr] (g1) -- (g2);
\end{tikzpicture}

\caption{$\textit{Decompose}(I)$ converts the instruction into ordered
sub-goals, each storing its completion type, target entities, spatial
relation or direction, and the top instruction-grounded visual query.}
\label{fig:decompose_example}

\end{figure*}
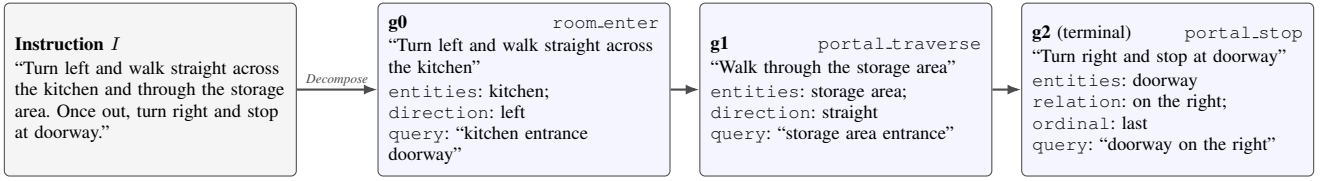

For each active sub-goal, \textit{SelectTarget} identifies a visual
target and \textit{ExecuteLocal} performs a motion segment.
After execution, \textit{BuildVisit} and \textit{Associate} build the anchor and update the graph, and \textit{VerifyProgress} determines whether the current stage can
advance. If verification fails, \textit{Explore} selects alternative directions
within the budget, while \textit{Backtrack} recovers previous states through
the anchor graph.

\begin{algorithm}[t]
\caption{Navigation Loop}
\label{alg:rgb_anchor_loop}
\begin{algorithmic}[1]
\REQUIRE instruction $I$, RGB observations
\STATE $G\leftarrow\textit{Decompose}(I)$; $M\leftarrow\textit{InitializeMemory}(o_0)$
\FOR{each sub-goal $g\in G$}
    \WHILE{$\textit{decision} \neq \texttt{Commit\_stage}$ } 
        \STATE $(v,p)\leftarrow\textit{SelectTarget}(o_t,g,M)$
        \STATE $(s_{\mathrm{exe}},o_{\mathrm{stop}},a_{\mathrm{local}})\leftarrow\textit{ExecuteLocal}(v,g)$
        \IF{$s_{\mathrm{exe}}=\texttt{Failure}$}
            \STATE $M\leftarrow\textit{UpdateFailure}(M,g)$
            \STATE $o_t\leftarrow\textit{Explore}(M,g)$ or $\textit{Backtrack}(M,g)$
        \ELSIF{$s_{\mathrm{exe}}=\texttt{Arrival}$} 
            \STATE $\textit{visit}\leftarrow\textit{BuildVisit}(o_{\mathrm{stop}},a_{\mathrm{local}})$
            \STATE $(\textit{state},M)\leftarrow\textit{Associate}(\textit{visit},M)$
            \STATE $\textit{decision}\leftarrow\textit{VerifyProgress}(g,\textit{state},\textit{visit},M)$
            \IF{$\textit{decision}=\texttt{Commit\_stage}$}
                \STATE move to next sub-goal
            \ELSE
                \STATE $o_t\leftarrow\textit{Explore}(M,g)$ or \textit{Backtrack}(M,g)
            \ENDIF
        \ENDIF
    \ENDWHILE
\ENDFOR
\end{algorithmic}
\end{algorithm}

Visual anchors link these modules by providing local control targets, place identities, and evidence for progress verification and recovery, organizing depth-free and pose-free navigation via graph memory. As Algorithm~\ref{alg:rgb_anchor_loop} defined, where $\textit{state}\in\{\texttt{Same}, \texttt{New},\texttt{Ambiguous}\}$, $\textit{decision}\in\{\texttt{Commit\_stage}, \texttt{Intermediate},\texttt{Wrong},\texttt{Uncertain}\}$, and $a_{\mathrm{local}}$ denote the local action sequence.

\subsection{Instruction Decomposition}
\label{subsec:method_decomposition}

At the beginning of an episode, the frozen LLM
implements $\textit{Decompose}(I)$ and converts the instruction into an ordered sequence
$G=\{g_1,\ldots,g_N\}$. Each sub-goal describes an independently observable event, such as reaching an object, entering a room, passing a doorway, or following a direction. It stores the completion type, target entities, etc. as shown in Fig.~\ref{fig:decompose_example}. Analogous to converting ambiguous queries into verifiable task constraints~\citep{chen2026geometrically}, this typed representation turns free-form language into explicit completion conditions that downstream modules can check, rather than letting the VLM implicitly decide when a stage ends. The controller advances to the next sub-goal only after \textit{VerifyProgress} returns \texttt{Commit\_Stage}.

\begin{figure}[t]
    \centering
    \includegraphics[width=0.85\linewidth]{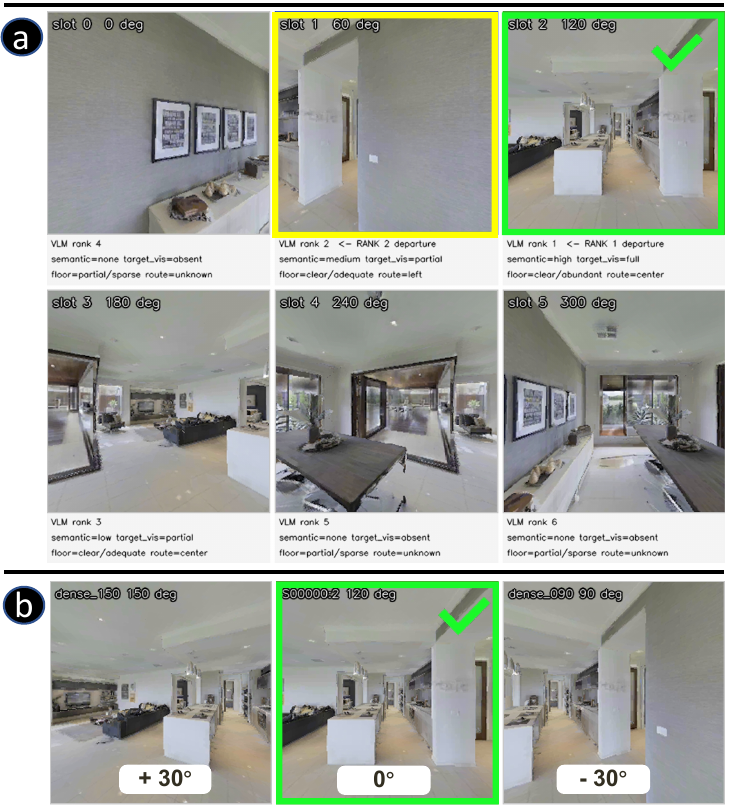}
    \caption{Visual target selection (\textit{SelectTarget}).
    (a) Six canonical views captured by rotating in place; the VLM ranks
    them with the sub-goal text and walkable-route evidence (green: Rank-1
    departure, yellow: Rank-2). (b) Refinement over three views at
    $\pm30^{\circ}$ around the coarse pick.}
    \label{fig:target_selection}

\end{figure}

\subsection{Visual Target Selection}
\label{subsec:method_target}

For an active sub-goal $g_i$, \textit{SelectTarget} converts language into an
executable visual target. The agent rotates in place and captures six
canonical RGB views separated by $60^\circ$. A VLM ranks these views using the sub-goal description, target entities, spatial relations, and traversability cues, then selects a candidate direction. Additional nearby views refine the selection and produce the target image $v$ for local execution.

Open-vocabulary detection and segmentation localize object instances in $v$;
a floor mask supplies traversable regions for doorways, corridors, and stairs.
The function outputs $(v,p)$, where $p$ is a trackable point in the selected
RGB observation. It is a visual reference, not a coordinate-defined waypoint.

\begin{figure*}[t]
    \centering
    \includegraphics[width=\textwidth]{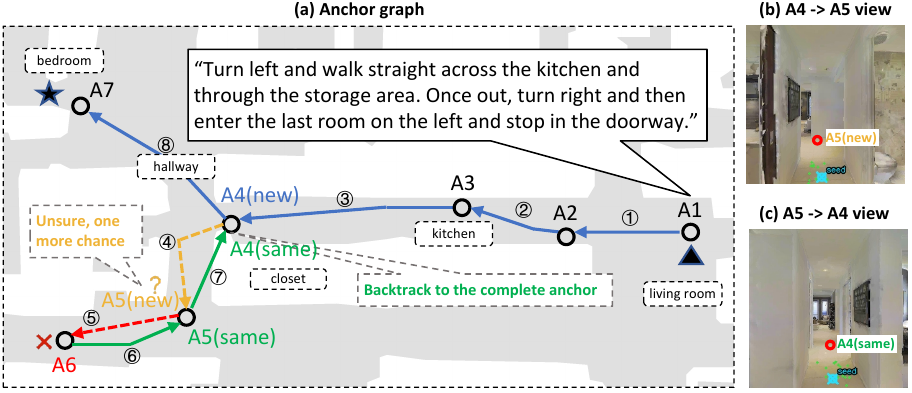}
    \caption{Coordinate-free anchor graph and visual backtracking.
    (a) Anchors $A_1$--$A_7$ are created by \textit{BuildVisit} from
    six-view RGB visits and linked by executed \texttt{TraversalEdge}s
    (solid blue, numbered in execution order); room labels are the semantic
    signatures used by \textit{Associate}. At $A_5$ the branch is judged
    uncertain (orange) and then fails (red). (b) The stored departure view
    of edge $A_4\!\rightarrow\!A_5$. (c) The view after \textit{Backtrack}
    returns along the same edge; \textit{Associate} classifies the returned
    visit as \texttt{SAME} to $A_4$ (green), which commits the reverse edge
    and restores exploration from $A_4$. The top-down map is for
    visualisation only and is never available to the policy.}
    \label{fig:anchor_backtrack}
    \vspace{-10pt}
\end{figure*}

\subsection{Local Execution}
\label{subsec:method_local}

\textit{ExecuteLocal}$(v,p,g)$ initializes a KLT~\citep{lucas1981iterative,tomasi1991detection,shi1994good} point tracker and gives a
crop around $p$ to the pretrained ViNT local navigator~\citep{shah2023vint}.
The tracker updates the point in each RGB frame and the goal crop follows it,
allowing ViNT to select forward and turning actions from the current visual
state. The module performs short-range continuous motion and does not plan the
long-horizon instruction. A point leaving the image boundary or being lost during tracking indicates failure of the current local target and returns execution signal $s_{\mathrm{exe}}=\texttt{Failure}$. 

\begin{figure}[t]
    \centering
    \includegraphics[width=\linewidth]{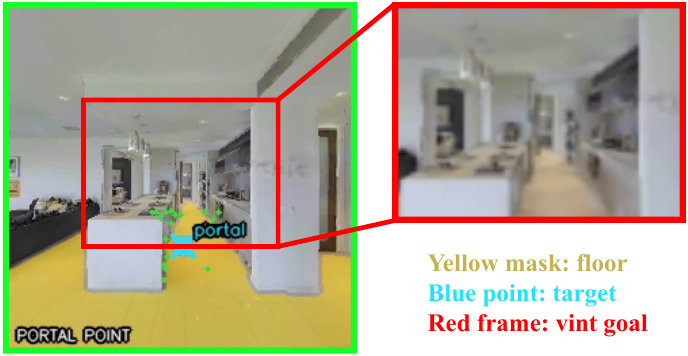}
    \caption{Goal construction for \textit{ExecuteLocal}. Left: the
    selected view $v$ with the floor mask (yellow) and the tracked target
    point $p$ (blue). Right: the goal crop around $p$ (red frame) that is
    fed to the ViNT local navigator; the tracker updates $p$ every frame
    and the crop follows it until $p$ exits through the bottom of the
    image.}
    \label{fig:vint_target}
\end{figure}

The local arrival candidate is triggered entirely by point-tracking geometry.
For object, floor, and route points, when a valid tracked point moves
downward and exits through the bottom boundary of the RGB observation
(``\texttt{bottom\_exit}''), the agent has arrived the selected local visual
target. \textit{ExecuteLocal} then returns $s_{\mathrm{exe}}=\texttt{Arrival}$,
the local action sequence $a_{\mathrm{local}}$, observation at stop $o_{\mathrm{stop}}$. VLM
inference, depth estimation, and simulator pose do not trigger this event.
It only indicates that local visual motion ended; task completion is decided
after the physical stop.

\subsection{Instruction Progress Verification}

\subsubsection{Anchor Construction}
\label{subsec:method_memory}

For observation at stop $o_{\mathrm{stop}}$, \textit{BuildVisit} captures a six-view RGB scan ($0^\circ,60^\circ,\ldots,300^\circ$) stored as a \textit{visit} containing views, sub-goal context, semantic descriptions, and motion evidence. Unseen places initialize a new \texttt{AnchorNode}, whereas revisited places merge into existing anchors storing a canonical visit, visual features, a semantic signature, and six relative direction slots. Executed motions and observation are logged as directed \texttt{TraversalEdge}s.

As shown in Fig.~\ref{fig:anchor_backtrack}(a), the resulting anchor graph maintains place identities and confirmed transitions without explicit coordinates; direction slots lack distance or global angles, and forward edges do not imply reverse paths. 

\textit{Associate}(\textit{visit}, \textit{M}) executes coarse-to-fine RGB place association via global descriptor retrieval with cyclic view alignment, refined by RANSAC~\citep{fischler1981random} local feature matching~\citep{rublee2011orb} and semantic verification. Outputs are \texttt{SAME}, \texttt{NEW}, or \texttt{AMBIGUOUS}; ambiguous visits are retained but cannot advance instructions or serve as backtracking targets.

\subsubsection{Progress Verification}
\label{subsec:method_progres}

As for \textit{VerifyProgress}$(g,\textit{state},\textit{visit},M)$, the verifier receives motion completion,
visit association, and the target entities, completion type, spatial relations,
and history of $g$. \textit{Associate} determines the visual place state,
while \textit{VerifyProgress} decides whether it satisfies the current
instruction stage.

If the completion condition is satisfied, the output is
\texttt{Commit\_stage}, and the controller advances $G$. A confirmed but
incomplete place is labeled \texttt{Intermediate} and stored as an anchor
without advancing the instruction. Incorrect and uncertain cases are classified as \texttt{Wrong} and \texttt{Uncertain}, and do not advance the task. Keeping uncertainty explicit prevents overconfident stage transitions from propagating over long horizons, a failure mode also observed in multi-turn tool-using agents~\citep{zhou2026exploring}; in contrast to training-based uncertainty alignment, Navi-Agent enforces it at inference time through the verification protocol.

\subsection{Backtrack and Relocalization}
\label{subsec:method_recovery}

When progress is not committed, \textit{CanExplore} checks whether the current
anchor has available exploration budget. If so, \textit{Explore} selects a new
direction and invokes \textit{SelectTarget} and \textit{ExecuteLocal} again.
Rejected directions are recorded in the anchor memory and are not repeatedly
explored for the same sub-goal. When the exploration budget is exhausted or
the current branch is judged inconsistent with the instruction, the agent
activates \textit{Backtrack}.

Unlike geometric backtracking, \textit{Backtrack} does not rely on pose or
metric maps. Each anchor transition stores historical RGB observations and
motion trajectories from local execution. During recovery, the agent retrieves
historical observations, matches the current view with the stored trajectory,
and recovers a visual target $(v',p')$ from the matched historical state.
The agent then reuses \textit{ExecuteLocal} with $(v',p')$ to iteratively return
to previous locations. If no reliable visual correspondence is found, the
recovery process terminates.

After each recovery step, \textit{BuildVisit} and \textit{Associate} verify the
returned location. Only a \texttt{Same} association with the target anchor is
accepted, while ambiguous or incorrect returns are rejected. After returning
to a confirmed anchor, failed directions remain stored in memory and
\textit{SelectTarget} chooses alternative candidates for the same sub-goal.

\begin{table*}[t]
\centering
\caption{Comparison with zero-shot VLN-CE methods on
\texttt{OpenNav\_R2R-CE\_100}. 
Y/N indicates whether the policy directly uses depth or coordinate/pose information.
View indicates the visual observation setting.}
\label{tab:main_results}
\setlength{\tabcolsep}{7pt}
\begin{tabular*}{0.95\textwidth}{@{\extracolsep{\fill}}cccccccc}
\hline
\textbf{Method} &
\textbf{Depth} &
\textbf{Coord/Pose} &
\textbf{View} &
\textbf{SR$\uparrow$} &
\textbf{OSR$\uparrow$} &
\textbf{NE$\downarrow$} &
\textbf{SPL$\uparrow$} \\
\hline
MapGPT-CE~\citep{chen2024mapgpt}
& Y & Y & Full Multi-view & 7.0 & 21.0 & 8.16 & 5.04 \\
DiscussNav~\citep{long2024discuss}
& Y & Y & Full Multi-view & 11.0 & 15.0 & 7.77 & 10.51 \\
InstructNav~\citep{long2024instructnav}
& Y & Y & Key Multi-view & 31.0 & -- & 6.89 & 24.00 \\
Open-Nav~\citep{qiao2025open}
& Y & Y & Full Multi-view & 16.0 & 23.0 & 7.25 & 12.90 \\
CA-Nav~\citep{chen2025constraint}
& Y & Y & Full Single-view & 25.3 & 48.0 & 7.58 & 10.80 \\
SmartWay~\citep{shi2025smartway}
& Y & Y & Full Multi-view & 29.0 & 51.0 & 7.01 & 22.46 \\
Fast-SmartWay~\citep{shi2025fast}
& Y & Y & Single-view & 27.8 & -- & 7.72 & 24.95 \\
\hline
DreamNav~\citep{wang2025dreamnav}
& Y & N & Key Multi-view & 32.8 & 41.0 & 7.06 & \textbf{28.95} \\
LightZeroNav~\citep{luo2026lightzeronav}
& N & N & Key Multi-view & 27.0 & 39.0 & 7.45 & 20.00 \\

\textbf{Navi-Agent (ours)}
& N & N & Key Multi-view &
\textbf{33.4} &
\textbf{48.3} &
\textbf{5.79} &
18.73 \\
\hline
\end{tabular*}
\vspace{-8pt}
\end{table*}

\section{Experiments}
\label{sec:experiments}

\subsection{Benchmark and Evaluation Protocol}
\label{subsec:experimental_setup}

\textbf{R2R-CE and Habitat.}
We evaluate Navi-Agent on the Habitat-based R2R-CE benchmark
\citep{savva2019habitat,chang2017matterport3d,anderson2018vision}. We use the corrected
100-episode \texttt{val\_unseen} subset from the OpenNav protocol~\citep{qiao2025open},
denoted as \texttt{OpenNav\_R2R-CE\_100}. Each episode provides a natural-language
instruction and an initial RGB observation. The agent interacts with Habitat
using four discrete actions: \texttt{move\_forward}, \texttt{turn\_left},
\texttt{turn\_right}, and \texttt{stop}. The forward step is $0.25$ m, each
turn is $10^\circ$, and each episode is limited to 500 actions.

\textbf{Information protocol.}
Navi-Agent only receives RGB observations, instructions, and online visual
memory. Depth, pose, odometry, SLAM/SfM coordinates, metric maps, and
shortest-path hints are unavailable during inference. Simulator pose is used
only for evaluator-side distance computation.

\textbf{Metrics.}
We report Success Rate (SR), Oracle Success Rate (OSR), and Navigation Error
(NE). SR requires the agent to stop within $3$ m geodesic distance of the goal, OSR measures whether the trajectory ever reaches this region, and NE denotes the final geodesic distance. SPL is reported when available.

\subsection{Main Results}
\label{subsec:main_results}

Table~\ref{tab:main_results} compares Navi-Agent with representative recent zero-shot VLN-CE systems, separating metric spatial state-based methods from geometry-constrained approaches. ``Global coord / Pose'' denotes a unified coordinate frame, and ``Depth'' indicates metric depth. Navi-Agent targets a stricter RGB-only setting: it uses neither metric spatial priors (e.g., pose or depth) nor global coordinates. Metric-based methods are included to highlight the benefit of additional spatial priors rather than for direct comparison. For methods without public implementations, we report the success rates from their original papers.

\subsection{Ablation Study}
\label{subsec:ablation_study}

We evaluate the effect of the semantic backbone, recovery mechanism, and
exploration policy while keeping the remaining modules and evaluation protocol
fixed.

\subsubsection{VLM Backbone}
\label{subsec:vlm_ablation}

We next vary the semantic backbone while keep other configuration fixed. This experiment
tests whether progress and place verification depend on one particular VLM.

\begin{table}[H]
\centering
\caption{VLM backbone ablation.}
\vspace{-5pt}
\label{tab:vlm_ablation}
\begin{tabular}{lcccc}
\hline
\textbf{Semantic model} & SR$\uparrow$ & OSR$\uparrow$ & NE$\downarrow$ & SPL$\uparrow$ \\
\hline
Qwen3.6-Plus(open-source)~\citep{qwen2026qwen36plus} & \textbf{33.4} & \textbf{48.3} & \textbf{5.79} & \textbf{18.73} \\
GPT-5.6-sol(close-source)~\citep{openai2026gpt56changelog} & 26.3 & 46.6 & 6.50 & 18.37 \\
GPT-5.5(close-source)~\citep{singh2025openai} & 28.6 & 47.3 & 6.24 & 16.82 \\
\hline
\end{tabular}
\end{table}

We will analyze both the absolute performance and the variance across models. All VLM backbones achieve competitive performance that meets or exceeds the baseline, with \textbf{Qwen3.6-Plus} achieving the best overall results. Beyond navigation, these performance margins suggest that our system could potentially be extended in the future as a benchmark to evaluate MLLMs in embodied AI tasks.

\subsubsection{Anchor Graph, Exploration and Recovery Strategy}
\label{subsec:exploration_recovery_ablation}

To evaluate the contribution of graph-based exploration and recovery, we ablate the exploration strategy while keeping the semantic reasoning and local navigation modules fixed. Specifically, the variant \textit{w/o anchor graph and recovery} disables the graph construction entirely; thus, it removes the distinction between exploration and verification, forcing the agent to move greedily until it actively stops or reaches the maximum step limit. The other variants enable backtracking via the anchor graph with varying maximum numbers of confirmed outward hops. In this benchmark, the full Navi-Agent uses a 1-hop exploration budget.

\begin{table}[H]
\centering
\caption{Exploration and recovery strategies.}
\vspace{-5pt}
\label{tab:exploration_recovery_ablation}

\begin{tabular}{lcccc}
\hline
\textbf{Configuration} &
\textbf{SR$\uparrow$} &
\textbf{OSR$\uparrow$} &
\textbf{NE$\downarrow$} &
\textbf{SPL$\uparrow$} 
\\
\hline
w/o anchor graph and recovery &
28.33 & 46.33 & 5.64 & \textbf{22.45}
\\
1 hop (Full Navi-Agent)&
\textbf{33.43} & \textbf{48.33} & \textbf{5.79} & 18.73
\\
2 hops &
21.67 & 40.33 & 6.58 & 14.39
\\
3 hops &
14.67 & 32.67 & 9.83 & 11.72
\\
\hline
\end{tabular}
\vspace{-5pt}
\end{table}

As shown in Table~\ref{tab:exploration_recovery_ablation}, removing recovery yields the highest SPL (22.45) by eliminating backtracking on successful paths, though at the cost of lower task completion (28.33\% SR). Conversely, increasing exploration depth beyond 1 hop severely degrades performance (SR drops to 14.67\% at 3 hops), as multi-hop recovery causes excessive graph expansion along wrong branches and compounds errors. Thus, a constrained 1-hop recovery achieves the optimal balance between failure correction and concise path execution.

\subsection{Spatial State Verification and Recovery}
\label{subsec:spatial_recovery}

\textbf{Spatial state verification.}
To independently evaluate the core backtracking capabilities, we assess whether the Visual Anchor Graph provides reliable spatial memory for visited places and transitions. Given a trajectory segment, the agent returns to a previous anchor using stored visual targets and motion histories without pose information. We measure the return position error and success rate, defined as $\mathbb{I}(d(p_{\mathrm{return}},p_{\mathrm{anchor}})<\delta)$. We further evaluate anchor association on true and visually similar revisitation pairs, where the agent predicts whether current and historical observations correspond to the same anchor, evaluated by accuracy, precision, recall, and F1 score.

As shown in Table~\ref{tab:spatial_verification}, Navi-Agent demonstrates strong physical backtracking and visual verification capabilities. Specifically, it achieves a low return position error of 2.32\,m with an 80.3\% physical revisitation success rate, proving that the Visual Anchor Graph provides reliable topological memory for precise backtracking. Furthermore, for visual place confirmation, our method obtains solid performance with F1 score as 0.71, confirming its solid capacity to disambiguate visually similar places and establish correct anchor associations.

\begin{table}[htb]
\centering
\caption{Spatial state verification and recovery evaluation.}
\vspace{-6pt}
\label{tab:spatial_verification}
\begin{tabular}{c}
\textbf{(a) Physical Revisitation}\\[2pt]
\begin{tabular}{lcc}
\hline
\textbf{Method} &
\textbf{Return Error $\downarrow$} &
\textbf{Return SR $\uparrow$}\\
\hline
Navi-Agent &
2.32 &
80.3\\
\hline
\end{tabular}
\\[8pt]
\textbf{(b) Visual Place Confirmation}\\[2pt]
\begin{tabular}{lcccc}
\hline
\textbf{Method} &
\textbf{Acc.} &
\textbf{Prec.} &
\textbf{Recall} &
\textbf{F1}\\
\hline
Navi-Agent &
0.71 &
0.68 &
0.75 &
0.71 \\
\hline
\end{tabular}

\end{tabular}
\vspace{-4pt}
\end{table}

\subsection{Real-World Deployment}
\label{subsec:real_world_deployment}

To demonstrate the generalization and robustness under strict sensor constraints, we deploy our system in a real-world indoor library with two platforms: a wheeled-legged robot~\citep{agibot2026g1} and an AGV. Both operate solely with a monocular RGB camera, without depth, IMU, or prior maps. The fixed control intervals yield an average step length of $0.5$\,m for forward motion and $\pm30^\circ$ for turning, with differences only in camera height and platform kinematics. We evaluate two target-directed tasks from the same room: Task 1 exits the room, turns right, and stops at a wooden door (10.5\,m), while Task 2 exits, follows the corridor, and stops near a green plant (21.5\,m).

As shown in Table~\ref{tab:real_world}, both platforms achieve consistent success rates of $0.7\text{--}0.8$, demonstrating strong cross-platform generalization. Initial failures occurred only during complex exits and turns; after a 1-hop exploration, our recovery module successfully backtracked to the correct topological node and completed navigation (up to $2/2$ recovery in long-horizon tasks). These results validate the adaptability and resilience of our mapless, depth-free framework under visual-only sensing.

\begin{table}[t]
\centering
\caption{Real-world deployment results (Succ.: successes out of 10 trials; Rec.: successful / triggered recoveries).}
\label{tab:real_world}
\vspace{-4pt}
\footnotesize
\setlength{\tabcolsep}{3pt}
\begin{tabular}{lccccc}
\hline
& & \multicolumn{2}{c}{\textbf{AGV}} & \multicolumn{2}{c}{\textbf{Wheeled-legged}} \\
\textbf{Task} & \textbf{Dist. (m)} & Succ. & Rec. & Succ. & Rec. \\
\hline
T1: exit $\rightarrow$ turn right $\rightarrow$ door & 10.5 & 8/10 & 2/2 & 8/10 & 1/1 \\
T2: exit $\rightarrow$ corridor $\rightarrow$ plant & 21.5 & 7/10 & 2/2 & 8/10 & 2/2 \\
\hline
\end{tabular}
\end{table}

\begin{figure}[t]
    \centering
    \includegraphics[width=0.162\linewidth, height=1.8cm]{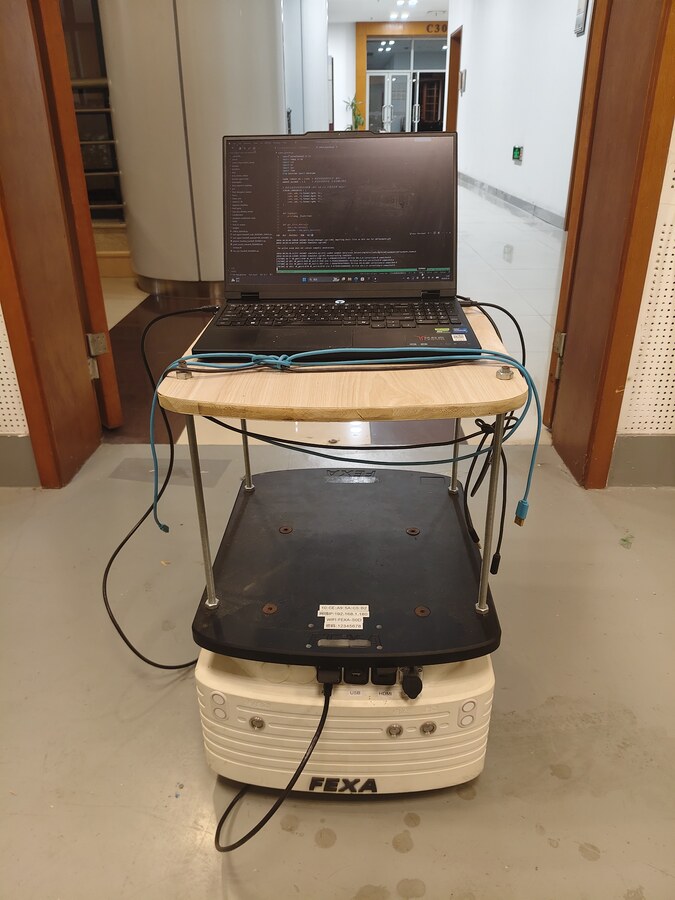}%
    \includegraphics[width=0.162\linewidth, height=1.8cm]{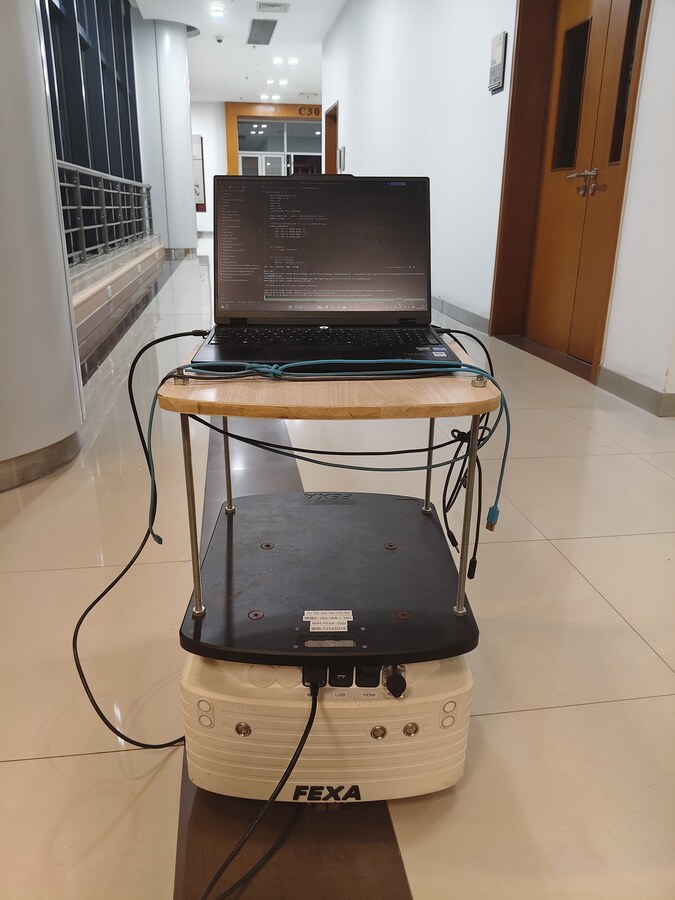}%
    \includegraphics[width=0.162\linewidth, height=1.8cm]{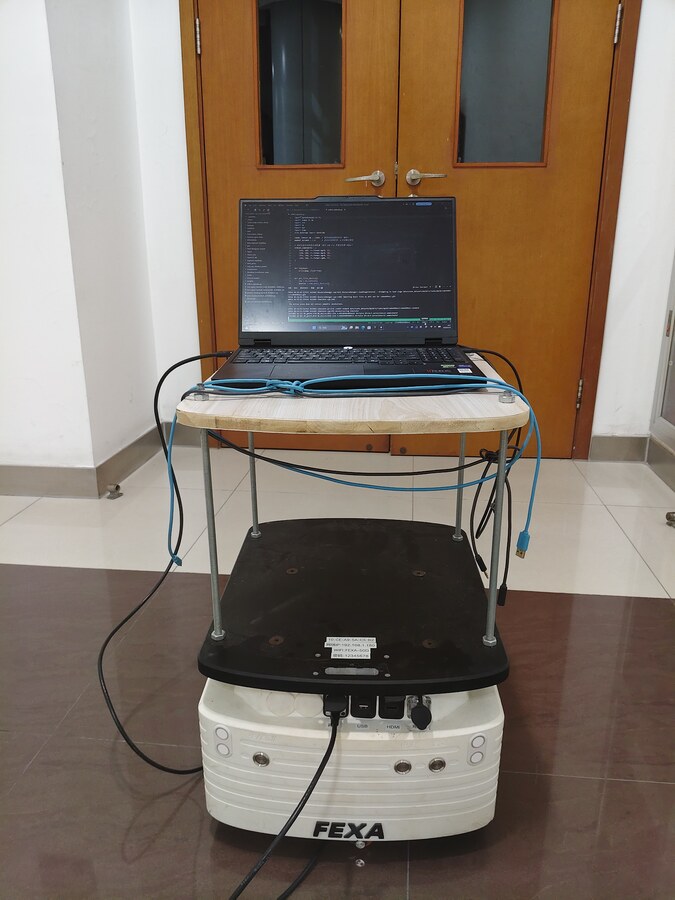}%
    \includegraphics[width=0.162\linewidth, height=1.8cm]{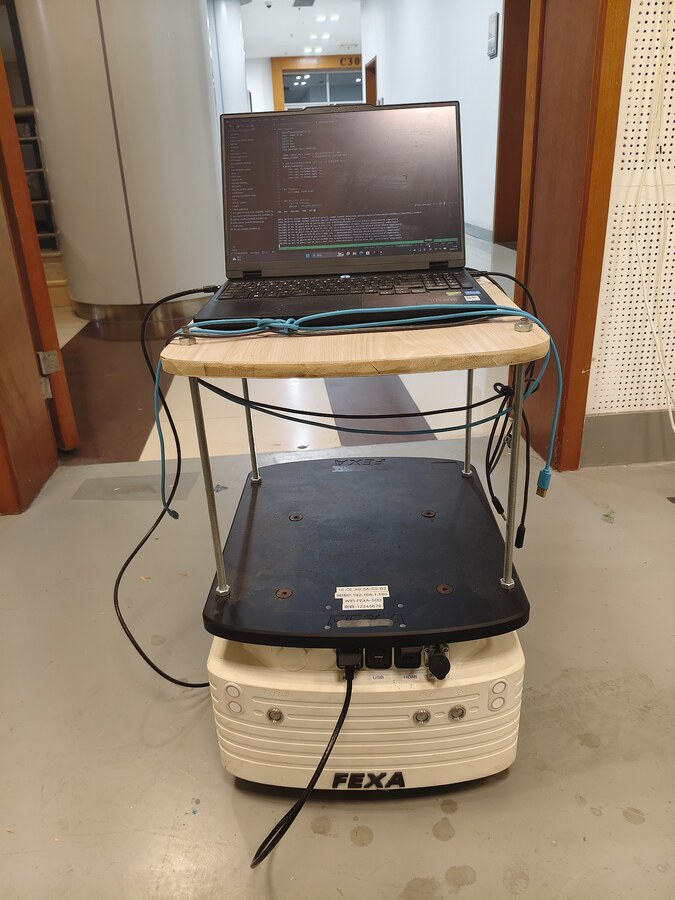}%
    \includegraphics[width=0.162\linewidth, height=1.8cm]{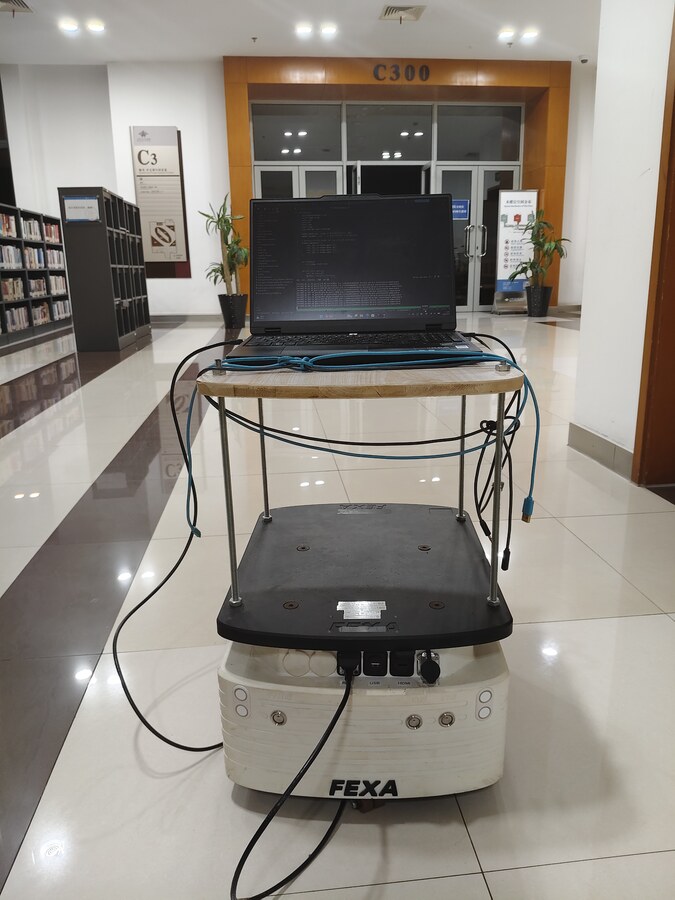}%
    \includegraphics[width=0.162\linewidth, height=1.8cm]{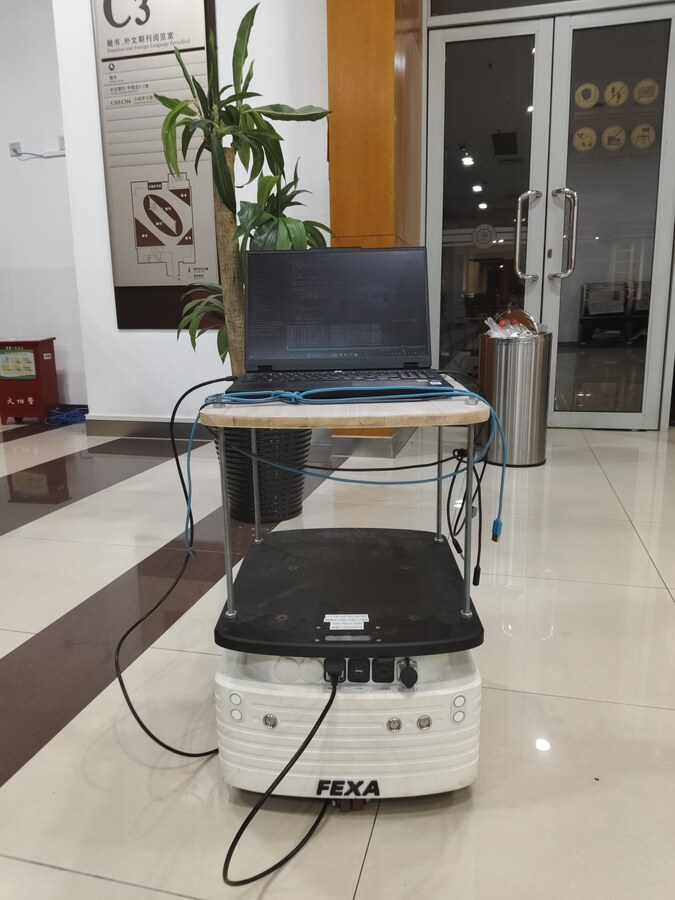}\\
    \includegraphics[width=0.162\linewidth, height=1.8cm]{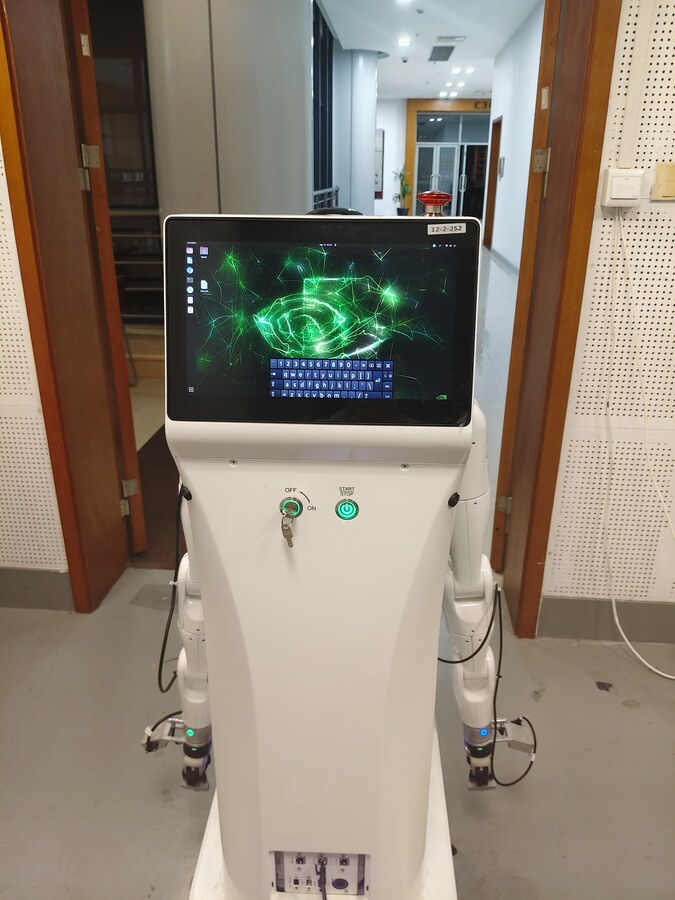}%
    \includegraphics[width=0.162\linewidth, height=1.8cm]{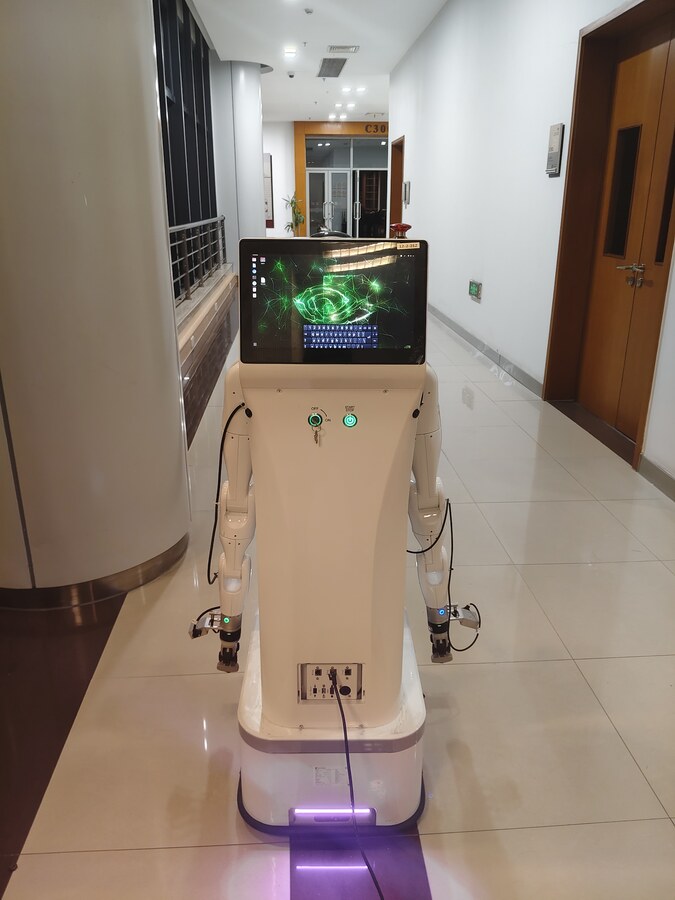}%
    \includegraphics[width=0.162\linewidth, height=1.8cm]{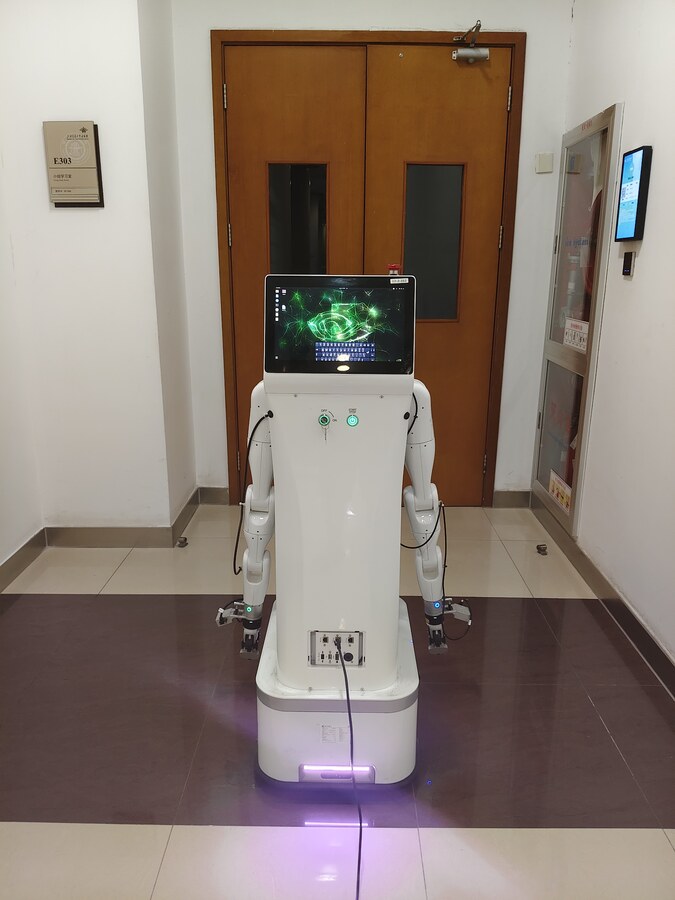}%
    \includegraphics[width=0.162\linewidth, height=1.8cm]{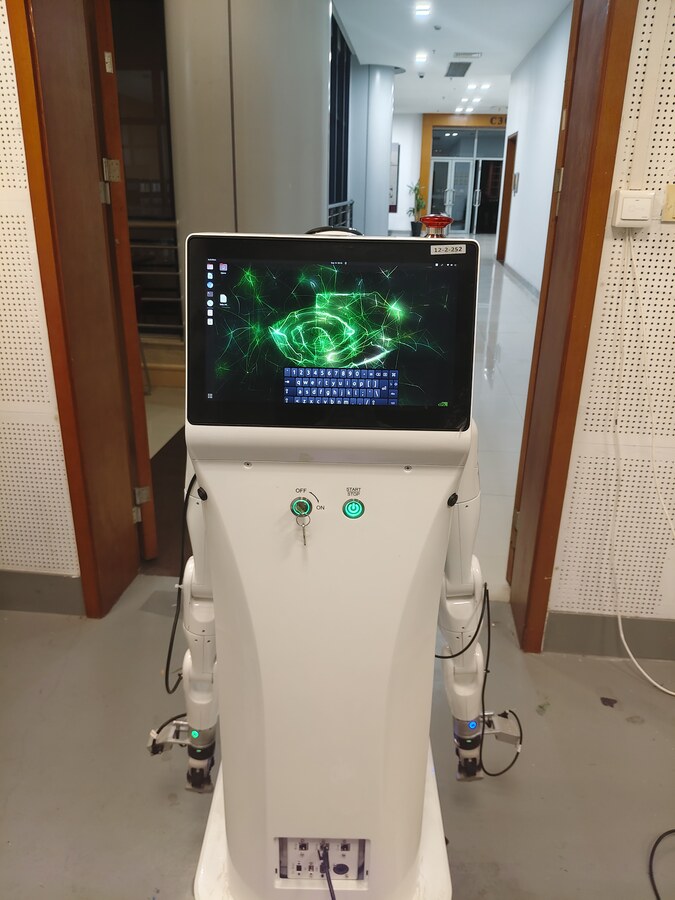}%
    \includegraphics[width=0.162\linewidth, height=1.8cm]{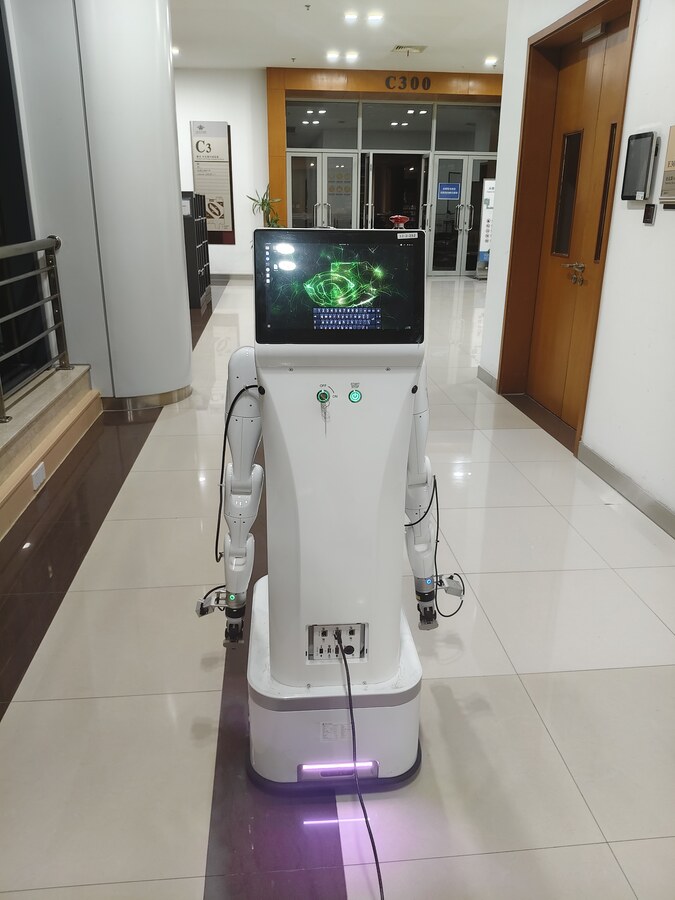}%
    \includegraphics[width=0.162\linewidth, height=1.8cm]{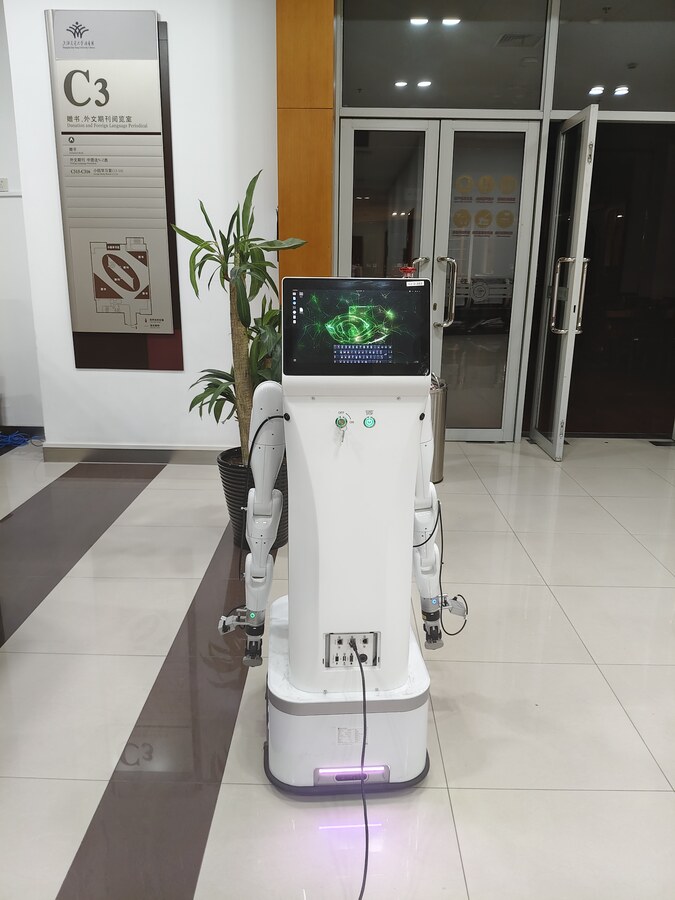}
    \vspace{-5pt}
    \caption{Real-world deployment of Navi-Agent across different platforms and tasks. The top row (1--6) shows execution traces on the AGV platform, while the bottom row (7--12) illustrates those on the wheeled-legged robot. Specifically, columns 1--3 and 7--9 depict Task 1, whereas columns 4--6 and 10--12 demonstrate Task 2.}
    \label{fig:real_world_deployment}
    \vspace{-5pt}
\end{figure}

\subsection{Latency Analysis}
\label{subsec:latency}

To analyze the time cost of whole system, we
measure the latency of each navigation cycle. The latency is decomposed into
five components: visual target selection, local execution, anchor construction,
progress verification, and backtracking. Results are averaged over navigation
episodes.

\begin{table}[htb]
\centering
\caption{Latency analysis of Navi-Agent.}
\label{tab:latency}
\footnotesize
\begin{tabular}{lc}
\hline
\textbf{Component} & \textbf{Latency (s)} \\
\hline
Visual Target Selection (VLM) & 8.3 \\
Local Execution (ViNT + Tracking) & 0.2 per step \\
Anchor Construction and Graph Update & 4.6 \\
Progress Verification & 3.7 \\
Backtracking and Visual Revisitation & sum of the above \\
\hline
\end{tabular}
\vspace{-5pt}
\end{table}


\section{Conclusion}
\label{sec:conclusion_and_limitations}

In this work, we study zero-shot Vision-and-Language Navigation in Continuous
Environments (VLN-CE) under geometry-constrained settings without depth
or global coordinates. Navi-Agent shows that a coordinate-free spatial
state built from visual observations and motion history supports long-horizon
navigation. By maintaining visual place identities and motion transitions,
it enables place confirmation, progress verification, and deviation recovery
via visual revisitation.

Extensive experiments on zero-shot VLN-CE benchmarks show Navi-Agent's
effectiveness. It achieves state-of-the-art results among geometry-constrained
methods and competes with approaches using geometric localization.
Analysis verifies that its spatial state supports reliable place association,
return verification, and exploration recovery. Real-world deployment on
different robots further proves its applicability.

In the future, we will extend visual spatial states
toward higher-level semantic spatial memory and reasoning. While current
work focuses on transitions for navigation, future
directions include inferring spatial relationships,
and reasoning over ambiguous configurations to build scalable spatial
understanding beyond explicit trajectories.




\footnotesize
\setlength{\bibsep}{0pt}
\bibliographystyle{IEEEtran}
\bibliography{references}
\end{document}